\documentclass[10pt,twocolumn,letterpaper]{article}

\usepackage[pagenumbers]{cvpr}      
\definecolor{cvprblue}{rgb}{0.21,0.49,0.74}
\usepackage[pagebackref,breaklinks,colorlinks,allcolors=cvprblue]{hyperref}
\usepackage{amsmath}
\usepackage{multirow}
\usepackage{colortbl}
\usepackage{wrapfig}
\usepackage{amsthm}
\usepackage[table]{xcolor} 
\usepackage{collcell}     
\usepackage{tikz}         
\def\paperID{4092} 
\def\confName{CVPR}
\def\confYear{2026}

\title{The Role of Entropy in Visual Grounding: Analysis and Optimization}

\author{\textbf{Shuo Li}\thanks{{ }Equal contributions.}$^{ \ \ ,1}$, \textbf{Jiajun Sun}$^{*,1}$, \textbf{Zhihao Zhang}$^{1}$,
\textbf{Xiaoran Fan}$^{1}$, \textbf{Senjie Jin}$^{1}$, \textbf{Hui Li}$^{1}$,  \textbf{Yuming Yang}$^{1}$,\\ \textbf{Junjie Ye}$^{1}$, \textbf{Lixing Shen}$^{2}$, \textbf{Tao Ji}$^{\dag,  1 }$, \textbf{Tao Gui}$^{ \dag, 1}$, \textbf{Qi Zhang}\thanks{{ }Corresponding author.}$^{ \ , 1}$, \textbf{Xuanjing Huang}$^{1}$\\
Fudan University$^{1}$  \  \ Hikvision Research Institute$^{2}$\\
\texttt{lis23@m.fudan.edu.cn, \{taoji, tgui, qz\}@fudan.edu.cn} 
}

\begin{document}
\maketitle
\begin{abstract}
Recent advances in fine-tuning multimodal large language models (MLLMs) using reinforcement learning have achieved remarkable progress, particularly with the introduction of various entropy control techniques. However, the role and characteristics of entropy in perception-oriented tasks like visual grounding, as well as effective strategies for controlling it, remain largely unexplored. To address this issue, we focus on the visual grounding task and analyze the role and characteristics of entropy in comparison to reasoning tasks. Building on these findings, we introduce ECVGPO (Entropy Control Visual Grounding Policy Optimization), an interpretable algorithm designed for effective entropy regulation. Through entropy control, the trade-off between exploration and exploitation is better balanced. Experiments show that ECVGPO achieves broad improvements across various benchmarks and models.
\end{abstract}    
\section{Introduction}
Large language models (LLMs), such as ChatGPT~\cite{ChatGPT}, have shown outstanding ability in handling a wide spectrum of text-based tasks. When equipped with visual encoders like CLIP~\cite{radford2021learningtransferablevisualmodels}, these models evolve into multimodal large language models (MLLMs) that extend their strengths to vision-related domains. This integration enables them to tackle various vision-language challenges, including image captioning~\cite{wang2020overview}, visual question answering~\cite{antol2015vqa}, and visual dialogue~\cite{das2017visual}.
Driven by the success of works like OpenAI's o1~\cite{openai2024openaio1card} and DeepSeek's R1~\cite{guo2025deepseek}, reinforcement learning finetuning of LLMs and MLLMs become a central focus for researchers. Notably, R1-like approaches often combine Reinforcement Learning with Verifiable Rewards (RLVR) and the Group Relative Policy Optimization (GRPO)~\cite{shao2024deepseekmath} algorithm. Such methods have proven particularly effective for complex reasoning tasks, including mathematical problem solving.

In reasoning tasks, entropy is an important metric for evaluating a model’s responses. Prior studies~\cite{wang20258020rulehighentropyminority,cheng2025reasoning} show that high-entropy tokens often cluster around specific types of words, such as transition terms (e.g., ``however'', ``but''), which play a critical role in the model’s reasoning process. Consequently, response entropy is frequently used as an indicator of a model’s exploration capability in reasoning tasks. However, as RLVR is applied to more non-reasoning tasks such as visual grounding~\cite{liu2025visual,shen2025vlm,fan2025grit}, whether entropy plays a different role has not yet been investigated. Studying entropy in visual grounding can provide crucial insights into model training, leading to more reliable and robust spatial localization.

In this paper, we analyze the behavior of entropy in a visual grounding task and identify a phenomenon that is opposite to that observed in reasoning tasks. For models trained on their respective reasoning tasks, entropy tends to drop to very low values after training. In contrast, in the visual grounding task, entropy remains high even after extensive training. Notably, these high-entropy tokens are primarily associated with the coordinate values of predicted answers and the object nouns that appear in the model’s reasoning process. Our investigation reveals that, during the GRPO training, the model consistently operates under higher entropy in visual grounding scenarios. We further analyze the cause of high entropy, observing that responses with a high advantage often correspond to low generation probabilities.

 Based on these findings, we propose Entropy Control Visual Grounding Policy Optimization (ECVGPO), a novel reinforcement learning method that employs self-information to quantify the confidence of the model’s generated responses and introduces a self-information penalty or encouragement in visual grounding training. By reshaping the advantages of positive samples, we leverage the generation probability of tokens (i.e., self-information) to increase or decrease their advantages, thereby encouraging the model to update more conservatively or aggressively. This enables explicit control over the model’s entropy, achieving a better balance between exploration and exploitation. 
 We evaluate our method on two of the most popular multimodal models: Qwen2.5-VL~\cite{bai2025qwen2} and InternVL3~\cite{zhu2025internvl3}. The experimental results show that ECVGPO improves the performance of the model on both the in-domain and out-of-domain evaluation benchmarks. At the same time, we conduct extensive ablation experiments to demonstrate the generalization capability of our proposed method.

In this paper, our main contributions are: 
\begin{itemize}[leftmargin=*]
    \item We compare the token-level entropy distributions between reasoning and visual grounding tasks, and find that models trained with RLVR on visual grounding exhibit consistently higher entropy, mainly because responses with high advantages often correspond to low probabilities;
    \item We propose ECVGPO, the first method that dynamically adjusts entropy during training to achieve a better trade-off between exploration and exploitation;
    \item We evaluate ECVGPO on the most popular MLLMs and demonstrate broad performance gains on benchmarks.
\end{itemize}

\section{Related Work}
\paragraph{Multimodal large language model}
Multimodal large language models (MLLMs), exemplified by GPT-4o~\cite{openai2024gpt4ocard}, are emerging as a central research focus in deep learning due to their impressive cross-domain capabilities. By integrating visual encoders with powerful language models, these systems enable joint visual–linguistic reasoning. Early milestones such as CLIP~\cite{radford2021learningtransferablevisualmodels} establish a strong link between textual and visual representations, laying the groundwork for numerous cross-modal applications. The BLIP series~\cite{li2022blip,blip2,dai2023instructblipgeneralpurposevisionlanguagemodels} extends this paradigm to tasks like visual question answering, whereas LLaVA~\cite{liu2024visual,liu2024llavanext} adopts a lightweight projection module and two-stage training to enhance spatial grounding and overall performance. More recent efforts, including MouSi~\cite{fan2024mousi} and Cambrian-1~\cite{tong2024cambrian1fullyopenvisioncentric}, leverage diverse vision encoders to improve multimodal comprehension. State-of-the-art models such as QwenVL~\cite{Qwen-VL,Qwen2-VL,Qwen2.5-VL}, InternLM-XComposer~\cite{internlmxcomposer,internlmxcomposer2}, and InternVL~\cite{chen2023internvl,chen2024far} follow design principles similar to LLaVA and consistently achieve leading results.
\paragraph{Reinforcement learning with verifiable reward}
Reinforcement Learning with Human Feedback (RLHF)~\cite{bai2022training} has become a cornerstone for aligning large language models (LLMs) with human intent. However, the quality and scalability of human feedback remain significant challenges. To address this, a new paradigm has emerged: Reinforcement Learning with Verifiable Rewards (RLVR)~\cite{shao2024deepseekmath}. Instead of relying on subjective human preferences, this method leverages a reward function that can be algorithmically verified. Pioneering works like O1~\cite{openai2024openaio1card} from OpenAI and R1~\cite{guo2025deepseek} from DeepSeek exemplify this direction, particularly in complex reasoning tasks such as mathematics~\cite{shao2024deepseekmath,yu2025dapo,chen2025bridging}. These methods often design reward functions that verify the correctness of each reasoning step or the final answer, allowing for more objective and scalable policy optimization. Recently, several studies, such as Visual-RFT~\cite{liu2025visual}, VLM-R1~\cite{shen2025vlmr1stablegeneralizabler1style}, UniVG-R1~\cite{bai2025univgr1reasoningguideduniversal} and ViGoRL~\cite{sarch2025groundedreinforcementlearningvisual}, introduce visual grounding tasks into RLVR frameworks to enhance perception-oriented capabilities.

\paragraph{Entropy's role in RLVR training}
Recent research has identified entropy as a crucial metric in RLVR. For example, \cite{wang20258020rulehighentropyminority} analyzes token-level entropy distributions in Chain-of-Thought (CoT)~\cite{wei2023chainofthoughtpromptingelicitsreasoning} reasoning and finds that only a small fraction of tokens exhibit high entropy, acting as pivotal decision points that guide the model toward diverse reasoning trajectories. Building on this, \cite{cui2025entropymechanismreinforcementlearning} reports an empirical relationship between entropy and downstream performance, indicating that policy performance is fundamentally constrained by the available policy entropy and reaches a predictable upper bound when entropy is fully exhausted. Similarly, ~\cite{cheng2025reasoningexplorationentropyperspective} revisits entropy as an exploration signal in reinforcement learning, examining its role in facilitating exploratory reasoning behaviors in large language models. While these studies~\cite{zhang2025edge,tan2025gtpo,su2025gppo} provide valuable insights into the role of entropy for complex reasoning tasks, its role in visual perception tasks—such as visual grounding—remains largely unexplored.

\begin{figure*}[ht!]
  \centering
   \includegraphics[width=0.88\linewidth]{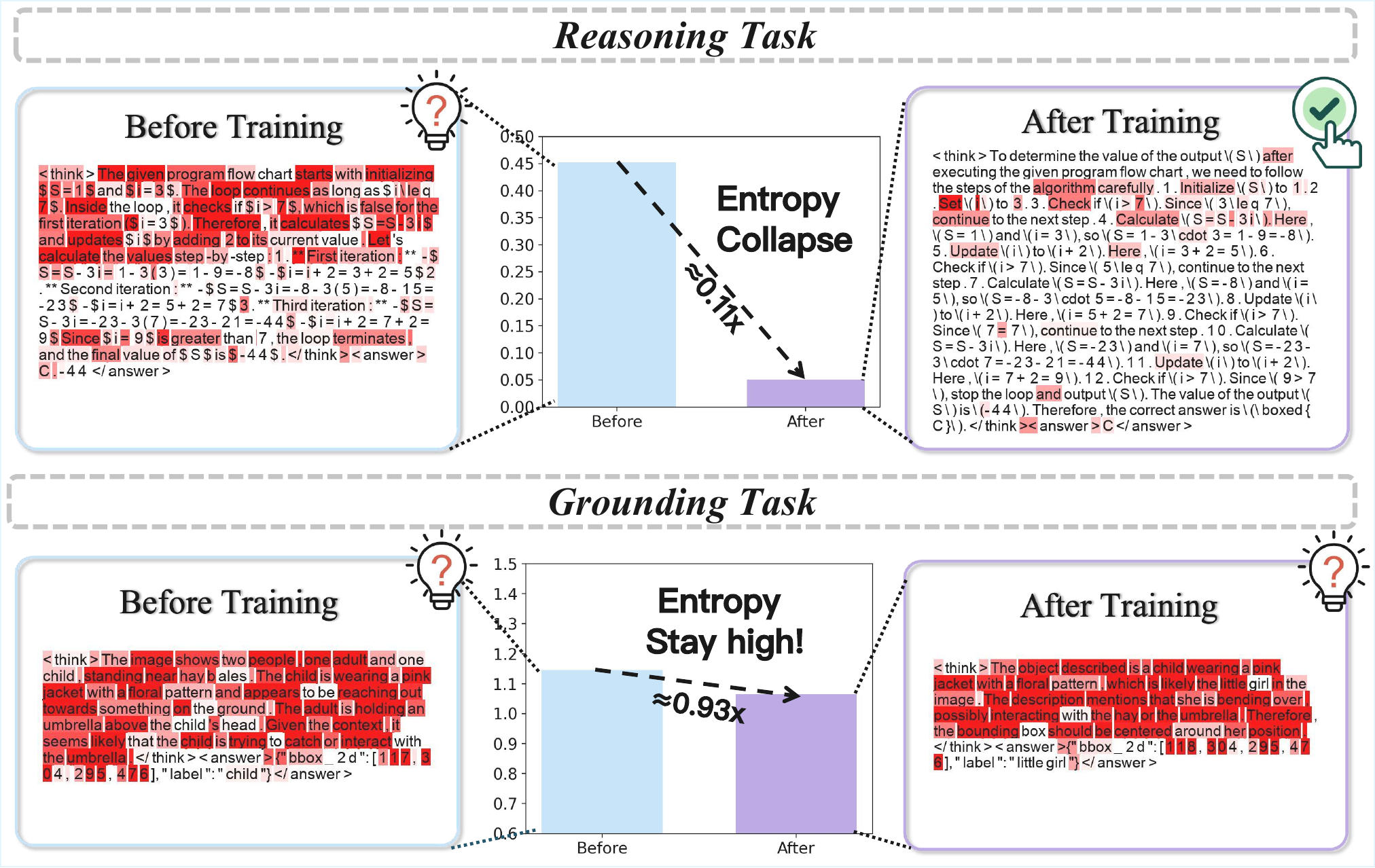}
   \caption{Comparison of entropy between reasoning and grounding tasks. Darker red indicates higher token entropy. In reasoning tasks, entropy rapidly decreases after training (entropy collapse), whereas in grounding tasks, it remains consistently high.}
   \label{fig:main}
\end{figure*}
\section{Preliminary}
\label{sec:3_pre}
In this section, we present the preliminaries, covering the training process of RLVR in the visual grounding task and highlighting the distinct roles that entropy plays in reasoning tasks versus visual grounding tasks.

\subsection{RLVR Training in Visual Grounding Task}

In the visual grounding setting, in line with most related work~\cite{shen2025vlmr1stablegeneralizabler1style,liu2025visualrftvisualreinforcementfinetuning}, we utilize Group Relative Policy Optimization (GRPO), which require a separate critic network to estimate policy values. GRPO directly compares multiple candidate responses from the same prompt, removing the dependency on an additional value function.

\paragraph{GRPO training}Given a query $q$, the policy $\pi_\theta$ samples $N$ candidate outputs $\{O_1, O_2, \ldots, O_N\}$, and each candidate is evaluated by a task-specific reward function $R(q, O_i)$. The relative quality is measured by computing the standardized advantage:
\begin{equation}\label{eq:define_advantage}
A_{i,t} = A_i= \frac{r_i - \text{mean}\{r_1, r_2, \ldots, r_N\}}{\text{std}\{r_1, r_2, \ldots, r_N\}},
\end{equation}
where $A_{i,t}$ represents the advantage of the token $o_{i,t}$ in response $O_i$ with respect to others in the same batch. The policy is optimized to increase the probability of responses with higher $A_i$ while constraining deviation from the reference model via a KL penalty:
\begin{equation}
\begin{split}
\mathcal{J}_{GRPO}(\theta) &= \mathbb{E}_{\{o_i\}_{i=1}^N \sim \pi_{\theta_{old}}(q)} \Bigg[ \frac{1}{N} \sum_{i=1}^N \big\{ \min[C_1, C_2] \\
&\quad - \beta \mathbb{D}_{KL}[\pi_\theta \| \pi_{ref}] \big\} \Bigg],
\end{split}
\end{equation}
\begin{align}
C_1 &= \frac{\pi_\theta(o_i|q)}{\pi_{\theta_{old}}(o_i|q)} \cdot A_i,\\
C_2 &= \text{clip}\left(\frac{\pi_\theta(o_i|q)}{\pi_{\theta_{old}}(o_i|q)}, 1+\epsilon, 1-\epsilon\right) \cdot A_i.
\end{align}

Here, $\epsilon$ is the clipping range, and $\beta$ controls the KL regularization strength. 

\paragraph{Reward setting}For the Visual Grounding task, consistent with~\cite{shen2025vlmr1stablegeneralizabler1style,liu2025visualrftvisualreinforcementfinetuning}, reward $R$ consists of two reward components:

\noindent\textbf{Accuracy reward}  
In the visual grounding task, the objective is to predict the bounding box of the target entity described in the input query.  
Let $\hat{b}$ be the predicted bounding box obtained from the model output $y$, and $b_{\mathrm{gt}}$ be the ground truth box.  
The accuracy reward is defined as:
\begin{equation}
\mathcal{R}_{\mathrm{acc}}(x, y) = \operatorname{IoU}\big(b_{\mathrm{gt}}, y\big),
\end{equation}
where $\operatorname{IoU}$ denotes the Intersection-over-Union score, which measures the overlap ratio between the predicted and reference boxes, thus incentivizing accurate localization.

\noindent\textbf{Format reward} This checks whether the output follows the required JSON-style structure within the \texttt{<answer>} tag and contains bounding box coordinates in the format \texttt{(<think>...</think><answer>\{...[x1, y1, x2, y2]...\}</answer>)}, assigning 1 if correct and 0 otherwise.

By jointly optimizing these rewards under the GRPO framework, the model learns to produce both accurate and well-formatted responses for visual grounding.

\subsection{High Entropy in Visual Grounding Task}
\paragraph{Token-level entropy}
We begin by defining the token-level entropy, which measures the uncertainty of the model’s output distribution at each decoding step. 
Formally, the token-level entropy at time step \( i \) is computed over the vocabulary distribution as:
\begin{align}\label{eq:entropy_token}
   \mathcal{H}_{o_i} = - \sum_{v \in \mathcal{V}} \pi_\theta(v|q, o_{<i}) \log \pi_\theta(v|q, o_{<i}),
\end{align}
where \( \mathcal{V} \) denotes the vocabulary, and \( \pi_\theta(v|q, o_{<i}) \) is the predicted probability of token \( v \) given the input \( q \) and the previously generated tokens \( o_{<i} \).

\paragraph{Policy entropy}Furthermore, we compute the policy entropy by definition, i.e., the average token-level entropy evaluated over the current policy model ${\pi}_\theta$ and the dataset $\mathcal{D}$:
\begin{align}
    \hat{\mathcal{H}}({\pi}_\theta,\mathcal{D}) = \mathbb{-E_{{\pi}_\theta,\mathcal{D}}}[\log \pi_\theta(o_i|q,o_{<i})]= \sum_{O \in \mathcal{D}}\sum_{o_i \in O} \mathcal{H}_{o_i}
\end{align}
\begin{figure}[t]
  \centering
   \includegraphics[width=0.95\linewidth]{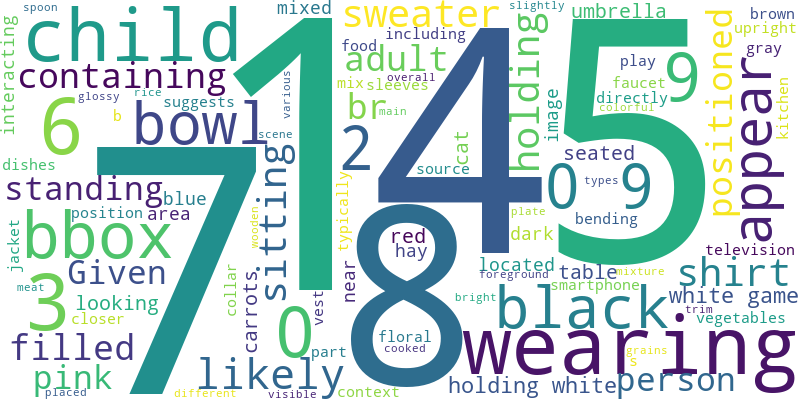}
   \caption{Word cloud of tokens with twice the average entropy.}
   \label{fig:wordcloud}
\end{figure}
We use LMM-R1~\cite{peng2025lmm}, trained on the reasoning task, and VLM-R1~\cite{shen2025vlmr1stablegeneralizabler1style}, trained on visual grounding, both based on Qwen2.5-VL-3B for comparison, and analyze the high-entropy tokens among the generated outputs, as shown in~\cref{fig:main}. The distribution of these tokens differs substantially between the two tasks, with the disparities primarily manifesting in two aspects:

\begin{enumerate}
    \item \textbf{Positional distribution.}  
    In reasoning tasks, high-entropy tokens appear more frequently during the reasoning process (i.e., between \texttt{<think>} and \texttt{</think>}) and are relatively sparse in the final answer segment (\texttt{<answer>...</answer>}). After training, the number of high-entropy tokens decreases substantially. In contrast, in vision-based tasks, high-entropy tokens are prevalent in both the reasoning and answer segments, both before and after training as shown in~\cref{fig:wordcloud}.

    \item \textbf{Token type characteristics.}  
    In reasoning tasks, prior to training, and consistent with the observations reported in~\cite{wang20258020rulehighentropyminority}, high-entropy tokens often correspond to ``pivot tokens'' that preserve semantic continuity within the reasoning chain. After training, all types of high-entropy tokens are greatly reduced. In visual grounding tasks, however, These tokens are mainly object nouns occurring in the thinking process and \textbf{coordinate values contained in the predicted bounding box of the answer}.
    
\end{enumerate}

\subsection{Explaining the Persistent Occurrence of High Entropy}\label{sec:why_high}
We aim to investigate one of the fundamental reasons behind the persistent occurrence of high-entropy tokens during visual grounding training.
\paragraph{Entropy change under policy gradient}
Following the analysis in \cite{cui2025entropymechanismreinforcementlearning} and Suppl. A, the change of entropy 
 $\mathcal{H}$ during training under policy gradient can be described by the following \cref{eq:entropy_diff}:
\begin{equation}\label{eq:entropy_diff}
\begin{split}
    \mathcal{H}(\pi_\theta^{k+1}|s)-&\mathcal{H}(\pi_\theta^{k}|s)
    \approx \\
    &-\eta \cdot \mathrm{Cov}_{a\sim \pi_\theta^{k}}(\log\pi_\theta^{k}(a|s),\pi_\theta^{k}(a|s)A(s,a)).
\end{split}
\end{equation}
Here, $s$ denotes the state, $\pi_{\theta}^k$ is the policy network at iteration $k$, $\eta$ represents the learning rate, and $\mathrm{Cov}$ denotes the covariance. According to this formula, we observe that \textbf{the action $a$ with both high (or low) probability and high (or low) advantage tends to decrease entropy, whereas the opposite conditions lead to increased entropy}.
\paragraph{Expected low probability of high-advantage responses} In the visual grounding task, the expected probability of high-dominance responses remains low due to two factors: (1) the presence of numerous responses with similarly high rewards, and (2) annotation noise, which blurs the distinction between optimal and near-optimal responses.
\begin{figure}[t]
  \centering
   \includegraphics[width=0.8\linewidth]{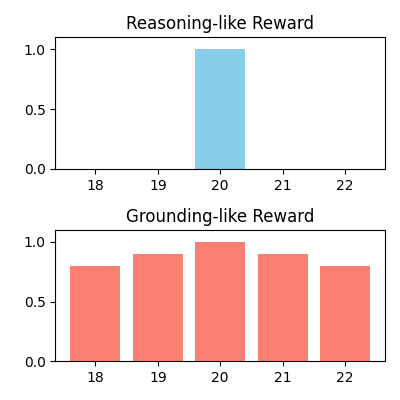}
   \caption{Comparison of the resoning-like reward and grounding-like reward.}
   \label{fig:diff_reward}
\end{figure}
Consider a simplified reward scenario as shown in
~\cref{fig:diff_reward}. In many \textbf{reasoning tasks}, the reward signal is sparse and binary. For a correct answer of $20$, the reward function $R_{\text{reasoning}}$ is defined as:
\begin{equation}
R_{\text{reasoning}}(a) =
\begin{cases}
1, & \text{if } a = 20, \\
0, & \text{otherwise}.
\end{cases}
\end{equation}
In this case, the policy gradient will concentrate probability mass on the single correct action, yielding a sharp, low-entropy distribution.

In contrast, consider a \textbf{grounding-like} task with a numerically continuous reward. 
For illustration, suppose the target value is $20$ and the reward is defined by a smooth decay function:
\begin{equation}
R_{\text{num}}(a) = \max\big(0, 1 - \lambda |a - 20|\big),
\end{equation}
where $\lambda > 0$ controls the decay rate. 
Here, the optimal action $a_{20} = 20$ receives the maximum reward $1$, but neighboring actions $a_{19}$ and $a_{21}$ still receive large rewards. Such a reward function encourages the policy to increase the probabilities of multiple adjacent actions with similar values.

Meanwhile, the answer coordinates generated in the visual grounding task often consist of approximately 10 tokens, and similar IoU scores can correspond to dozens or even hundreds of candidate answers. Due to the discrete nature of the tokens, these answers can be approximately viewed as different actions. That is, there is such a advantage set:
\begin{equation}
\begin{split}
&\{A_{\delta}\} = \{\, A(s,a) \mid |A - A_{\max}| \le \delta \},\\
&where \ |\{A_\delta\}| = M,\ M>>1.
\end{split}
\end{equation}
where \( \{A_{\delta}\} \) represents the advantage set of suboptimal solutions, and \( M \) denotes the total number of such suboptimal solutions.

\begin{figure}[t]
  \centering
   \includegraphics[width=0.8\linewidth]{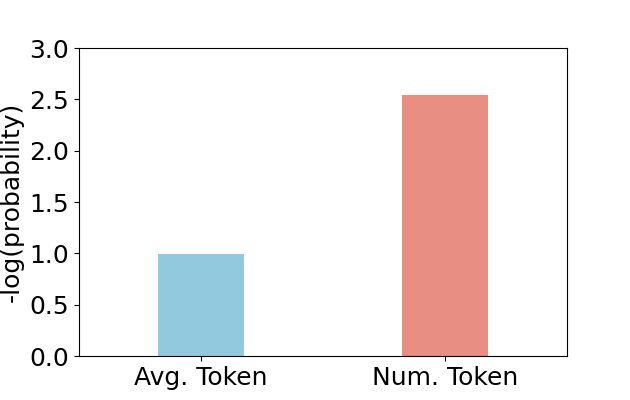}
   \caption{Comparison of the average probability of all tokens and numeric tokens in positive advantage samples during VLM-R1's training.}
   \label{fig:fig_diff_entropy}
\end{figure}

In addition, according to the conclusion of~\cite{chen2024revisitingreferringexpressioncomprehension}, due to the ambiguous annotation standards in the visual grounding task (e.g., an answer with IoU = 0.9 may be nearly identical to one with IoU = 0.95), it lacks the certainty present in most reasoning tasks, where the correct answer is typically a single numerical value or discrete option. Consequently, the expected advantage and probability of the optimal solution in the annotations and that of suboptimal solutions differ only marginally:
\begin{align}
\mathbb{E}({A_{\delta}})\ &\approx \mathbb{E}(A_{max}),\\
\mathbb{E}({\pi (s,a_{\delta})})\ &\approx \mathbb{E}(\pi (s,a_{max})) \leq \frac{1}{M}.
\end{align}
Therefore, in each training round, \textbf{sampled responses with high advantage $A_{high}$ often have low probabilities $\pi(a|s)$.} According to \cref{eq:entropy_diff}, this phenomenon intuitively results in a larger covariance, which in turn leads to higher entropy. The experimental results in~\cref{fig:fig_diff_entropy} confirm this phenomenon. 
Obviously and intuitively, when the entropy of the answer is high, numerous high-entropy tokens also appear in the thinking process. In addition, we offer an empirical discussion of underlying causes in the Suppl. C.


Overall, the behavior of entropy in visual grounding differs markedly from that in reasoning tasks. In visual grounding, a substantial number of high-entropy tokens persist, indicating that the model consistently operates under high uncertainty and low confidence when generating responses. This means that the entropy in the visual grounding task behaves differently; therefore, it is necessary to design specific strategies to regulate it for a better balance between exploration and exploitation.

\section{Method: ECVGPO}
Inspired by the findings in~\cref{sec:3_pre}, we introduce the \textbf{Entropy Control Visual Grounding Policy Optimization (ECVGPO)} algorithm for visual grounding. The key idea is to adjust the entropy by adjusting the covariance of probability and advantage at training time. Unlike reasoning tasks, which only aim to increase entropy to facilitate exploration, we consider reducing entropy when it is too high, improving the model's ability to exploit good policies. By attenuating policy updates in such cases, ECVGPO suppresses the reinforcement of unreliable patterns, thereby fostering the acquisition of more robust and generalizable representations.

We first compute the self-information \( S_i \) and  \( S_{i,t} \) that generated the 
\( i \)-th response and \( t \)-th response in \( i \)-th response to measure their uncertainty:
\begin{align}
    S_i &= -\frac{1}{|O|} \sum_{o_i \in O} \log \pi_\theta(o_i|q,o_{<i}),\\
    S_{i,t} &= - \log\pi_\theta(o_i|q,o_{<i}).
\end{align}

According to the definition of self-information, higher self-information corresponds to lower confidence in the output response, and vice versa~\cite{fu2025deepthinkconfidence}.

The standard deviation of the rewards for the sampled responses is computed as follow:
\begin{align}
    r_{std} = \sqrt{\sum_{i=1}^{|O|}{(r_i-\frac{1}{|O|}\sum_{i=1}^{|O|}r_i)^2}}.
\end{align}
A high reward combined with a low standard deviation indicates that the model has limited room for further policy updates.

After that, we reshape the advantages of \( i \)-th positive samples as follows:
\begin{align}\label{eq:new_advantage}
A_i' =
\begin{cases}
\max(\frac{A_{i,t}}{l_0},A_{i,t}-\frac{S_{i,t}}{r_0}), \text{when} \ A_i > 0 \ \& \ r_{std} < {\delta}\\
A_i, \text{when} \ A_i <0 
\end{cases},
\end{align}
Here, \( A_i \) denotes the advantage of the \( i \)-th sample, as obtained in \cref{eq:define_advantage}. The coefficient \( l_0 > 1\) is the lower-bound clipping factor, which is always greater than 0 to ensure that the advantage remains positive after the reshape, thereby preventing a positive example from being transformed into a negative one. The coefficient \( |r_0| > 1\) is used to control the change in entropy. By enforcing \(r_{std} < {\delta}\), we restrict the intervention to situations where the training is stable, preventing instability caused by premature adjustments.



Finally, we obtain the new training objective:
\begin{equation}
\begin{split}
\mathcal{J}_{ECVGPO}(\theta) &= \mathbb{E}_{\{o_i\}_{i=1}^N \sim \pi_{\theta_{old}}(q)} \Bigg[ \frac{1}{N} \sum_{i=1}^N \big\{ \min[C_1, C_2] \\
&\quad - \beta \mathbb{D}_{KL}[\pi_\theta \| \pi_{ref}] \big\} \Bigg],
\end{split}
\end{equation}
\begin{align}
C_1 &= \frac{\pi_\theta(o_i|q)}{\pi_{\theta_{old}}(o_i|q)} \cdot A_i',\\
C_2 &= \text{clip}\left(\frac{\pi_\theta(o_i|q)}{\pi_{\theta_{old}}(o_i|q)}, 1+\epsilon, 1-\epsilon\right) \cdot A_i'.
\end{align}

In addition, ECVGPO offers strong interpretability. Specifically, after applying reshape to advantages, the covariance term in~\cref{eq:entropy_diff} can be approximately as follows:
\begin{align}
    \mathrm{Cov}_{a\sim \pi_\theta^{k}}(\log\pi_\theta^{k}(a|s),\pi_\theta^{k}(a|s)[A(s,a)-\textcolor{red}{k \cdot \log\pi_\theta^{k}(a|s)}]).
\end{align}
Where $k$ is a coefficient. According to the analysis in~\cref{sec:why_high}, the advantage of positive samples is typically large, while their generation probability tends to be small. Therefore, when $k>0$, the covariance between advantage and probability decreases, leading to a reduction in the policy model’s entropy. Conversely, when $k<0$, the covariance between the two increases, thereby increasing the model’s entropy. Compared with some entropy regularization methods in reasoning tasks, our method involves fewer interventions, since visual grounding naturally exhibits higher entropy and less training stability.


\section{Experiment}
\subsection{Setup}
\paragraph{Implementation detail}
We use the training splits of RefCOCO, RefCOCO+, and RefCOCOg~\cite{kazemzadeh2014referitgame,yu2016modelingcontextreferringexpressions} as our training data, which are the most widely used datasets for the visual grounding task. We randomly selected a total of 5.6k samples from these three datasets.
Our codebase is based on VLM-R1~\cite{shen2025vlmr1stablegeneralizabler1style}. Specifically, we set rollout = 8, temperature = 1.0, number of iterations = 1, KL divergence ratio (i.e., $\beta$) = 0.04, learning rate = $1\times10^{-6}$, batch size = 4, gradient accumulation = 1. The experiments are conducted with 8$\times$H100 GPUs. See Suppl. B for more Settings.
\paragraph{Baseline}
We choose three state-of-the-art models, \textbf{Qwen2.5-VL-3B-Instruct}, \textbf{InternVL3-2B-Instruct} and \textbf{InternVL3-8B-Instruct}, as our base models. 
We compare our proposed method with three variants: the base model, the Supervised Fine-Tuning (SFT) model, and the GRPO-based fine-tuning model. 
For a fair comparison, the base model and SFT-based model are trained and evaluated in the same format as the base model's standard documentation. GRPO-based method uses the same training data as ECVGPO, and the training settings for GRPO are kept consistent with those of ECVGPO.
\paragraph{Benchmark}
For evaluation, we use the validation split of RefCOCO/+/g~\cite{kazemzadeh2014referitgame,yu2016modelingcontextreferringexpressions} as the in-domain benchmark, while the test splits of LISA-Grounding~\cite{lai2024lisareasoningsegmentationlarge} and RefGTA~\cite{tanaka2019generatingeasytounderstandreferringexpressions} serve as out-of-domain benchmarks.
As a reasoning-centric benchmark, LISA-Grounding requires models to handle complex tasks such as detailed visual analysis, understanding nuanced referring expressions, and performing relational reasoning among objects to correctly determine the target region. Meanwhile, RefGTA is constructed from game scenes rather than the general scenarios in RefCOCO. We adopt the same subset used in VLM-R1~\cite{shen2025vlmr1stablegeneralizabler1style}.
These two out-of-domain test sets are employed to evaluate the generalization ability of the model under different perception and reasoning scenarios.

\subsection{Main Result}
\paragraph{Comparison of ECVGPO with baseline methods}
\newcommand*{\MinNumber}{-2.5} 
\newcommand*{\MaxNumber}{1.5}  
\definecolor{MinColor}{rgb}{0.110, 0.329, 0.659} 
\definecolor{MaxColor}{rgb}{0.933, 0.176, 0.165}

\definecolor{MidColor}{rgb}{1.0, 1.0, 1.0} 

\newcommand{\ColorCell}[2]{%
    
    \pgfmathsetmacro{\inputValue}{#1} 

    
    \pgfmathsetmacro{\numbervalue}{min(\inputValue, \MaxNumber)}
    \pgfmathsetmacro{\numbervalue}{max(\numbervalue, \MinNumber)}

    
    \ifdim\numbervalue pt < 0pt 
        \pgfmathsetmacro{\Percent}{max(0, 100 * \numbervalue / \MinNumber)}
        \xdef\currentcolor{MinColor!\Percent!MidColor} 
    \else 
        \pgfmathsetmacro{\Percent}{max(0, 100 * \numbervalue / \MaxNumber)}
        \xdef\currentcolor{MaxColor!\Percent!MidColor}
    \fi

    \cellcolor{\currentcolor} #2
}

\begin{table*}[!ht]
  \centering
  \scalebox{0.76}{
  \setlength\tabcolsep{4pt}
\begin{tabular}{l|cccccccc|cc|c}
\toprule
  \multirow{3}{*}{\textbf{Method}} & \multicolumn{8}{c|}{\textbf{In-Domain}} & \multicolumn{2}{c|}{\textbf{Out-of-Domain}}  & \multirow{3}{*}{\textbf{Overall}}\\
\cmidrule(lr){2-9} \cmidrule(lr){10-11} 
 & RefCOCO & RefCOCO & RefCOCO & RefCOCO+ & RefCOCO+ & RefCOCO+ & RefCOCOg & RefCOCOg & LISA & RefGTA & \\
 & \small{(val)} & \small{(testA)} & \small{(testB)} & \small{(val)} & \small{(testA)} & \small{(testB)} & \small{(val)} & \small{(test)} & \small{(test)} & \small{(test)} &  \\
\midrule
\rowcolor[gray]{0.9}\multicolumn{12}{c}{\textbf{Qwen2.5-VL-3B-Instruct}} \\
Base Model&  88.05&  91.55&  83.20&  79.35&  86.65&  72.35&  84.80&  84.40&  56.82&    72.45& 79.96\\
w/ SFT &  88.75&  91.85&  84.80&  80.35&  87.40&  74.15&  85.15&  85.25&    42.10& 69.75& 78.95\\
w/ GRPO & 90.16$_{\pm0.18}$ & 93.00$_{\pm0.43}$ & 86.16$_{\pm0.41}$ & 83.59$_{\pm0.45}$ & \textbf{89.34}$_{\pm0.42}$ & 76.53$_{\pm0.48}$ & 86.51$_{\pm0.35}$ & 86.29$_{\pm0.55}$ &  \textbf{65.90}$_{\pm1.14}$  & 73.90$_{\pm1.00}$& 83.13\\
w/ Ours &  \textbf{90.55}$_{\pm0.29}$&  \textbf{93.23}$_{\pm0.31}$&  \textbf{86.8}$_{\pm0.35}$&  \textbf{84.00}$_{\pm0.37}$&  89.13$_{\pm0.37}$&  \textbf{76.91}$_{\pm0.41}$&  \textbf{87.16}$_{\pm0.25}$&  \textbf{86.45}$_{\pm0.16}$&  65.16$_{\pm0.68}$&\textbf{74.42}$_{\pm0.65}$& \textbf{83.38}\\

\multicolumn{1}{c|}{$\Delta$ GRPO} & \ColorCell{0.39}{+0.39} & \ColorCell{0.22}{+0.22} & \ColorCell{0.64}{+0.64} & \ColorCell{0.41}{+0.41} & \ColorCell{-0.21}{-0.21} & \ColorCell{0.39}{+0.39} & \ColorCell{0.65}{+0.65} & \ColorCell{0.17}{+0.17} &  \ColorCell{-0.74}{-0.74}  & \ColorCell{0.53}{+0.53} & \ColorCell{0.25}{+0.25} \\ 

\rowcolor[gray]{0.9}\multicolumn{12}{c}{\textbf{InternVL3-2B-Instruct}} \\
Base Model&  \textbf{89.80}&  \textbf{93.20} &  \textbf{86.80}&  \textbf{83.45}& \textbf{88.95} & \textbf{76.25} &  \textbf{88.15}& 86.00 &  0.13$^{**}$&  26.50 & 71.92   \\
w/ SFT &  86.65&  91.20&  84.55&  78.9&  85.8&  70.95&  85.5&  84.3&  48.91& 13.50& 73.02\\
w/ GRPO &  89.21$_{\pm0.35}$&  92.58$_{\pm0.28}$&  86.62$_{\pm0.33}$&  81.95$_{\pm0.72}$&  88.82$_{\pm0.49}$&  74.67$_{\pm0.55}$&  87.50$_{\pm0.32}$&  \textbf{86.87}$_{\pm0.38}$&  51.12$_{\pm5.63}$& 51.26$_{\pm 2.48}$& 79.05\\
w/ Ours &  89.49$_{\pm0.51}$&  92.73$_{\pm0.17}$&  86.50$_{\pm0.17}$&  81.80$_{\pm0.40}$&  88.76$_{\pm0.25}$&  74.70$_{\pm0.49}$& 87.56$_{\pm0.32}$ & 86.80$_{\pm0.25}$& \textbf{59.26}$_{\pm1.76}$&    \textbf{53.49}$_{\pm1.85}$& \textbf{80.11}\\
\multicolumn{1}{c|}{$\Delta$ GRPO} & \ColorCell{0.28}{+0.28} & \ColorCell{0.15}{+0.15} & \ColorCell{-0.12}{-0.12} & \ColorCell{-0.15}{-0.15} & \ColorCell{-0.06}{-0.06} & \ColorCell{0.03}{+0.03} & \ColorCell{0.06}{+0.06} & \ColorCell{-0.07}{-0.07} & \ColorCell{8.14}{+8.14} & \ColorCell{2.33}{+2.33} & \ColorCell{1.05}{+1.06}\\

\rowcolor[gray]{0.9}\multicolumn{12}{c}{\textbf{InternVL3-8B-Instruct}} \\
Base Model&  \textbf{93.00}& \textbf{95.05} & \textbf{88.65} & \textbf{88.40} &  92.05& \textbf{81.35} & 89.35 & 90.05 & 0.84$^{**}$ &  28.65  & 74.73\\
w/ SFT &  88.3& 91.05 & 85.05 & 81.65 & 88.60 & 75.05 & 86.20 & 87.05 &  58.38  & 33.05& 77.43\\
w/ GRPO &  91.65$_{\pm0.65}$&  94.29$_{\pm0.44}$ & 88.01$_{\pm0.31}$& 87.33$_{\pm0.52}$ & 91.96$_{\pm0.60 }$&  80.51$_{\pm0.90}$& 88.97$_{\pm0.46}$ & 89.86$_{\pm0.52}$ &  78.78$_{\pm1.45}$  & 48.8$_{\pm6.30}$& 84.01\\
w/ Ours & 92.15$_{\pm0.15}$ & 94.42$_{\pm0.28}$ & 88.53$_{\pm0.13}$ & 88.13$_{\pm0.33}$ & \textbf{92.36}$_{\pm0.52}$ & 81.18$_{\pm0.71}$ &\textbf{89.43}$_{\pm0.41}$& \textbf{90.06}$_{\pm0.26}$ & \textbf{81.14}$_{\pm0.45}$ &  \textbf{51.29}$_{\pm6.24}$  &\textbf{84.87} \\
\multicolumn{1}{c|}{$\Delta$ GRPO} & \ColorCell{0.50}{+0.50} & \ColorCell{0.13}{+0.13} & \ColorCell{0.52}{+0.52} & \ColorCell{0.80}{+0.80} & \ColorCell{0.40}{+0.40} & \ColorCell{0.68}{+0.68} & \ColorCell{0.44}{+0.44} & \ColorCell{0.20}{+0.20} & \ColorCell{2.36}{+2.36} & \ColorCell{2.49}{+2.49} & \ColorCell{0.85}{+0.85}\\

\bottomrule
\end{tabular}
  }
    \caption{Compare results of EC-GRPO with other baseline methods on in-domain and out-of-domain benchmarks. We set $r_0=10, 25, -50$ for each model. For a fair comparison, the question-answering templates used in the base and SFT models follow the official templates. Considering the high variance of GRPO-type methods on the visual grounding task, we report the mean and standard deviation over four runs. Overall represents the average score of the in-domain and out-of-domain performance. $**$ indicates that the model does not follow the format correctly. The best performances within each metric are \textbf{bolded}.}
      \label{tab:main_result}
\end{table*}

\cref{tab:main_result} presents the comparison between our proposed ECVGPO and other baseline methods, including SFT and GRPO, across both in-domain and out-of-domain benchmarks on Qwen2.5-VL-3B, InternVL3-2B, and InternVL3-8B. To ensure fairness, all models adopt the official question-answering templates used in the base and SFT versions. Considering the high variance of GRPO-type methods in visual grounding tasks, we report both the mean and standard deviation over four independent runs for each configuration.

As shown in the table, ECVGPO consistently achieves superior performance compared with both GRPO and SFT across all model scales and datasets. On Qwen2.5-VL-3B, our method reaches an overall score of 83.38, outperforming GRPO (83.13) and clearly surpassing the SFT baseline (78.95). The improvement becomes more pronounced for larger models: ECVGPO yields a +1.06 gain on InternVL3-2B and +0.85 on InternVL3-8B, respectively. These results demonstrate that the proposed entropy-controlled mechanism enables more stable and effective policy optimization, leading to consistent benefits over both supervised and GRPO baselines.

In terms of stability, ECVGPO shows smaller standard deviations across most benchmarks, indicating improved training consistency. This confirms that our method effectively mitigates the high-variance issue often observed in GRPO-type algorithms by adaptively controlling entropy to maintain a balanced exploration–exploitation trade-off.


Overall, ECVGPO achieves both higher accuracy and greater stability than existing baselines, proving its effectiveness and scalability for MLLMs.

\begin{figure}[!ht]
  \centering
   \includegraphics[width=0.98\linewidth]{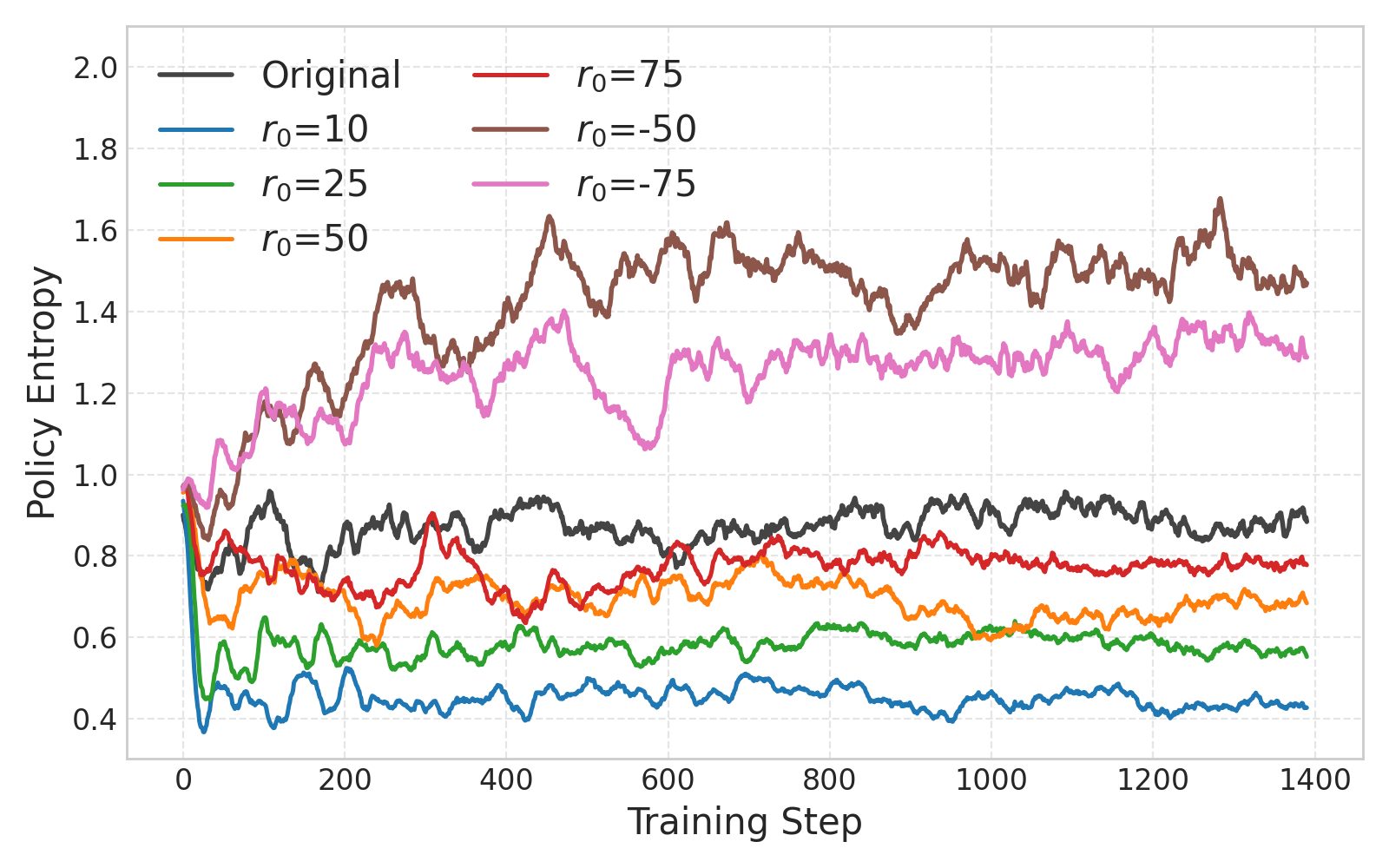}
   \caption{Trend of entropy under different parameters $r_0$. The experiments are conducted on the InternVL3-8B model. The lines in the figure are smoothed.}
   \label{fig:entropy}
\end{figure}

\paragraph{Comparison of training efficiency}
\begin{figure}[t]
  \centering
   \includegraphics[width=0.95\linewidth]{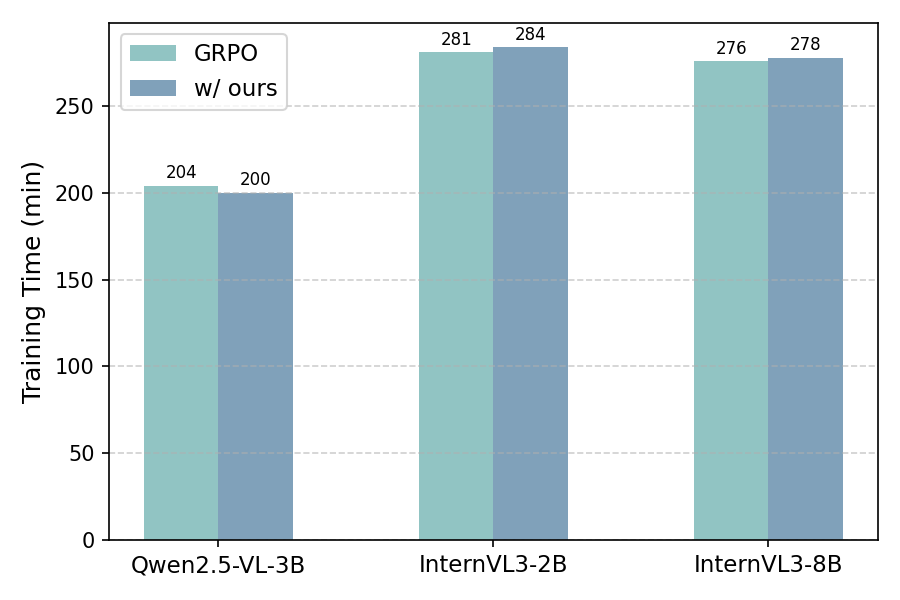}
   \caption{Comparison of the training time of GRPO and our method.}
   \label{fig:effi}
\end{figure}
As shown in~\cref{fig:effi}, the training time of our method is comparable to that of GRPO across all models. Specifically, for Qwen2.5-VL-3B, our method slightly reduces the training time (200 min vs. 204 min), while for InternVL3-2B and InternVL3-8B, the difference is marginal (within ±3 minutes). This indicates that incorporating our proposed optimization introduces almost no additional computational overhead, maintaining a similar level of training efficiency as the baseline GRPO. Therefore, our method maintains both effectiveness and efficiency, making it suitable for MLLMs' reinforcement fine-tuning.
\paragraph{Comparison of entropy trends}
Figure~\ref{fig:entropy} illustrates the trend of policy entropy under different values of $r_0$ on the InternVL3-8B model. 
The results show that the change in policy entropy is fully consistent with the trend of $1 / r_0$. 
Specifically, smaller (positive) values of $r_0$ correspond to lower entropy, while larger (negative) values of $r_0$ lead to higher entropy. 
This inverse relationship demonstrates that our method can effectively and smoothly control the entropy of the policy. 
Such controllability and consistency indicate that the proposed entropy control mechanism is both interpretable and stable, 
allowing the model to dynamically balance exploration and exploitation during training.
\subsection{Ablation Study}
\begin{table}[!ht]
\centering
  \setlength\tabcolsep{5pt}

  \scalebox{0.78}{
\begin{tabular}{lcccc}
    \toprule
    Setting  & RefCOCO$_{val}$      &RefCOCO+$_{val}$ & RefCOCOg$_{val}$ & LISA \\
    \midrule
      GRPO & 91.65$_{\pm0.65}$ & 87.33$_{\pm0.52}$ & 88.97$_{\pm0.46}$  & 78.78$_{\pm1.45}$   \\
    \midrule
     $r_0$=-50   & \textbf{92.15}$_{\pm0.15}$   & \textbf{88.13}$_{\pm0.33}$ & \textbf{89.43}$_{\pm0.41}$   & \textbf{81.14}$_{\pm0.45}$\\
    
    $r_0$=50           & 91.56$_{\pm0.43}$& 87.91$_{\pm0.47}$ & 88.88$_{\pm0.16}$    & 79.55$_{\pm0.19}$\\
    $r_0$=75         & 91.58$_{\pm0.22}$ & 87.43$_{\pm0.25}$   & 89.40$_{\pm0.60}$&80.71$_{\pm0.76}$ \\

 \hline
\end{tabular}
}
    \caption{Result of different $r_0$. The best performance is \textbf{bolded}. The experiments are conducted on the InternVL-8B model.We report the mean and standard deviation over four runs.
}
      \label{tab:r_0_result}
\end{table}
\paragraph{Comparison of different coefficients}
Table~\ref{tab:r_0_result} presents the performance of ECVGPO under different $r_0$ values on the InternVL3-8B model. 
Overall, the results demonstrate that the choice of $r_0$ has a clear influence on model performance. 
When $r_0 = -50$, the model achieves the best results across all benchmarks, surpassing the GRPO baseline by a large margin 
(+0.5 on RefCOCO, +0.8 on RefCOCO+, +0.5 on RefCOCOg, and +2.4 on LISA). 
This suggests that a moderate negative $r_0$ provides a proper level of entropy increase, 
enhancing exploration while maintaining stable optimization. 
When $r_0$ becomes positive (e.g., $r_0=50$ or $r_0=75$), the entropy is suppressed, 
leading to slightly worse performance due to insufficient exploration. 
These results validate that controlling entropy through $r_0$ effectively regulates the trade-off between exploration and exploitation, 
and that setting $r_0$ to a small negative value yields the best balance.
\paragraph{Comparison with other entropy-regularization methods}
\begin{table}[!ht]
\centering
  \setlength\tabcolsep{5pt}

  \scalebox{0.74}{
\begin{tabular}{lcccc}
    \toprule
    Setting  & RefCOCO$_{val}$      &RefCOCO+$_{val}$ & RefCOCOg$_{val}$ & LISA \\
    \midrule
      GRPO & 91.65$_{\pm0.65}$ & 87.33$_{\pm0.52}$ & 88.97$_{\pm0.46}$  & 78.78$_{\pm1.45}$   \\
    \midrule
     w/ En. Re.   & 1.25   & 1.15 & 1.5  & 0.72 \\
     
     w/ En. Adv.   & \textbf{92.40}$_{\pm0.40}$   & 87.68$_{\pm0.94}$ & 89.21$_{\pm0.37}$   & 78.90$_{\pm1.05}$\\    
     
     w/ Ours   & 92.15$_{\pm0.15}$   & \textbf{88.13}$_{\pm0.33}$ & \textbf{89.43}$_{\pm0.41}$   & \textbf{81.14}$_{\pm0.45}$\\

 \hline
\end{tabular}
}
    \caption{Result of different entropy regularization method. The best performance is \textbf{bolded}. The experiments are conducted on the InternVL-8B model. We report the mean and standard deviation over four runs.
}
      \label{tab:entropy_result}
\end{table}
As shown in~\cref{tab:entropy_result}, we compare different entropy regularization strategies under the InternVL-8B model. It can be observed that directly applying the entropy-based methods commonly used in reasoning-oriented reinforcement learning tasks (e.g., Entropy Reward~\cite{bai2022training,yu2022you} and Entropy Advantage~\cite{cheng2025reasoning}) is not well suited for the visual grounding task. In particular, the Entropy Reward method leads to severe training instability and performance collapse, reflected by extremely low scores across all benchmarks. This phenomenon indicates that simply increasing policy entropy fails to provide meaningful exploration in grounding scenarios, where visual information plays a dominant role.
In contrast, our proposed method achieves consistently strong and stable results across all datasets, outperforming the GRPO baseline and the Entropy Advantage method. Notably, our approach improves performance on out-of-domain datasets such as LISA, demonstrating both its effectiveness and robustness. These results validate that our entropy regularization design better balances exploration and stability in visual grounding.

\paragraph{More result}
Due to space limitations, additional ablation results are provided in the Suppl. D. 
\section{Conclusion}
In this work, we present a comprehensive analysis of entropy behavior in perception-oriented reinforcement learning, focusing on the visual grounding task. Our study reveals that entropy exhibits distinct dynamics compared to reasoning tasks, where high-advantage responses often correspond to low-generation probabilities, leading to persistently high entropy. Based on these insights, we propose ECVGPO, a simple yet effective extension of GRPO that reshapes advantages according to self-information to achieve more stable and interpretable optimization. Extensive experiments on multiple multimodal large language models and benchmarks demonstrate that ECVGPO consistently improves grounding accuracy, stability, and generalization without introducing additional training costs. We hope our findings provide a new perspective on understanding entropy in multimodal reinforcement learning and inspire further research into perception-oriented policy optimization.

{
    \small
    \bibliographystyle{ieeenat_fullname}
    \bibliography{main}
}
\appendix
\clearpage
\setcounter{page}{1}
\maketitlesupplementary


\section{Proof of Policy Entropy Change Approximation}
\label{sec:appendix:entropy_proof}

This section provides the detailed proof for the approximation of the policy entropy difference between two consecutive training steps, assuming a \textbf{Tabular Softmax Policy} updated via \textbf{Vanilla Policy Gradient (VPG)} with a learning rate $\eta$. The proof process is derived from~\cite{Liu2025RLPolicyEntropy} and~\cite{cui2025entropymechanismreinforcementlearning}.

\medskip
\noindent\textbf{Theorem 1 (Entropy Change under Policy Gradient)}
\textit{
Assuming the actor policy $\pi_{\theta}$ is a tabular Softmax policy updated by VPG, the difference in policy entropy at state $s$ between two consecutive steps is approximated by:
\begin{equation}
\begin{split}
    \label{eq:entropy_approximation}
    \mathcal{H}&(\pi_\theta^{k+1}|s)-\mathcal{H}(\pi_\theta^{k}|s)\approx \\ 
    &-\eta\cdot \mathrm{Cov}_{a\sim\pi_{\theta}^{k}(\cdot|s)}(\log~\pi_{\theta}^{k}(a|s),\pi_{\theta}^{k}(a|s)\cdot A(s,a))
\end{split}
\end{equation}
}
\medskip

\noindent\textbf{Proof} The proof proceeds in three main steps: 1) expressing the entropy difference in terms of logit changes, 2) deriving the logit changes under the VPG update rule, and 3) combining the results.

\subsection*{Step 1: Entropy Difference based on Logit Change}
For a Softmax policy $\pi_{\theta}(a|s)$, the change in policy entropy between steps $k$ and $k+1$ can be approximated using a first-order Taylor expansion (or a known result for Softmax policies) in terms of the logit difference $z_{s,a}^{k+1}-z_{s,a}^{k}$:
\begin{equation}
\begin{split}
    \label{eq:entropy_cov_logit}
    \mathcal{H}(\pi_{\theta}^{k+1}|s)&-\mathcal{H}(\pi_{\theta}^{k}|s)\approx \\ - &\mathrm{Cov}_{a\sim\pi_{\theta}^{k}(\cdot|s)}\left(\log\pi_{\theta}^{k}(a|s), z_{s,a}^{k+1}-z_{s,a}^{k}\right)
\end{split}
\end{equation}

\subsection*{Step 2: Logit Change under VPG Update}
For a tabular Softmax policy, the logit parameter $z_{s,a} = \theta_{s,a}$ is updated by VPG using the gradient 
$\nabla_{\theta_{s,a}} J(\theta)$:
$$z_{s,a}^{k+1}-z_{s,a}^{k} = \eta \cdot \nabla_{\theta_{s,a}} J(\theta)$$
The policy gradient at state $s$ is:
\begin{equation}
    \nabla_{\theta_{s,a}} J(\theta) = \mathbb{E}_{a'\sim\pi_{\theta}(\cdot|s)}\left[\nabla_{\theta_{s,a}}\log\pi_{\theta}(a'|s) \cdot A(s, a')\right]
\end{equation}
Using the identity for the gradient of the log-probability of a Softmax function, $\frac{\partial \log \pi_{\theta}(a'|s)}{\partial \theta_{s,a}} = \mathbb{I}_{a'=a} - \pi_{\theta}(a|s)$, we have:
\begin{equation}
\begin{split}
    \nabla_{\theta_{s,a}} J(\theta) &= \sum_{a'\in\mathcal{A}}\left[\pi_{\theta}(a'|s) \cdot (\mathbb{I}_{a'=a} - \pi_{\theta}(a|s)) \cdot A(s, a')\right] \\
    &= \pi_{\theta}(a|s) \cdot \left[ A(s, a) - \sum_{a'\in\mathcal{A}}\pi_{\theta}(a'|s) \cdot A(s, a') \right] 
\end{split}
\end{equation}

Under the standard VPG assumption that the advantage function is centered, $\sum_{a'}\pi_{\theta}(a'|s) \cdot A(s, a') = \mathbb{E}_{a'\sim\pi_{\theta}(\cdot|s)}[A(s, a')] = 0$, the gradient simplifies to:
$$\nabla_{\theta_{s,a}} J(\theta) = \pi_{\theta}(a|s) \cdot A(s, a)$$
Substituting this back into the logit difference equation yields:
\begin{equation}
    \label{eq:logit_diff_final}
    z_{s,a}^{k+1}-z_{s,a}^{k} = \eta\cdot\pi_{\theta}^{k}(a|s)\cdot A(s,a)
\end{equation}

\subsection*{Step 3: Final Substitution and Conclusion}
Substituting~\cref{eq:logit_diff_final} into the entropy approximation in~\cref{eq:entropy_cov_logit}:
\begin{equation}
    \begin{split}
      \mathcal{H}(\pi_{\theta}^{k+1}|s)&-\mathcal{H}(\pi_{\theta}^{k}|s)\approx\\ - &\mathrm{Cov}_{a\sim\pi_{\theta}^{k}(\cdot|s)}\left(\log\pi_{\theta}^{k}(a|s), \eta\cdot\pi_{\theta}^{k}(a|s)\cdot A(s,a)\right)  
    \end{split}
\end{equation}
Factoring the constant learning rate $\eta$ out of the covariance, we obtain the final result:
\begin{equation}
    \begin{split}
        \mathcal{H}(\pi_{\theta}&^{k+1}|s)-\mathcal{H}(\pi_{\theta}^{k}|s)\approx \\ &-\eta\cdot \mathrm{Cov}_{a\sim\pi_{\theta}^{k}(\cdot|s)}(\log~\pi_{\theta}^{k}(a|s),\pi_{\theta}^{k}(a|s)\cdot A(s,a))
    \end{split}
\end{equation}
\section{More Training Setting}
\begin{table}[ht]
\centering
\begin{tabular}{l| c c}
\toprule
Hyperparameter & QwenVL & InternVL \\
\midrule
Precision & \multicolumn{2}{c}{bf16} \\
Optimizer & \multicolumn{2}{c}{AdamW} \\
Image resolution & anyres & 1344*896(max) \\
Attention type & \multicolumn{2}{c}{flash attention} \\
Deepspeed stage &3 &2 \\
\(\delta\) &0.1 &\(\infty\)\\
\(l_0\) & \multicolumn{2}{c}{25} \\
\bottomrule
\end{tabular}
\caption{
Hyperparameters of our Qwen2.5-VL's and InternVL3's training.
}
\vspace{2mm}
\label{tab:training_setup}
\end{table}
\Cref{tab:training_setup} presents a detailed training setup.
\section{More Discussion for Low Probability Tokens}\label{sec:appendix:why_low}
During the thinking process, we also observe a large number of low-probability tokens. We attribute this phenomenon to two main qualitative reasons:

\paragraph{Mismatch between thinking steps and rewards} In reasoning tasks, the model typically performs step-by-step thinking, where each intermediate step explicitly contributes to the final answer generation. In contrast, for visual grounding tasks, only the final coordinate prediction receives a reward, making it difficult to determine which parts of the reasoning process actually influence the final output.

\paragraph{Limited in-domain performance improvement
} After GRPO training, the model (especially the InternVL3 series) shows little improvement over the base model on in-domain data. This suggests that the model struggles to learn a stable and effective thinking strategy under such training settings, which may further lead to the frequent occurrence of low-probability tokens.

\section{More Ablation Study}
\paragraph{Detailed training analysis}
\begin{figure*}[ht!]
  \centering
   \includegraphics[width=1.00\linewidth]{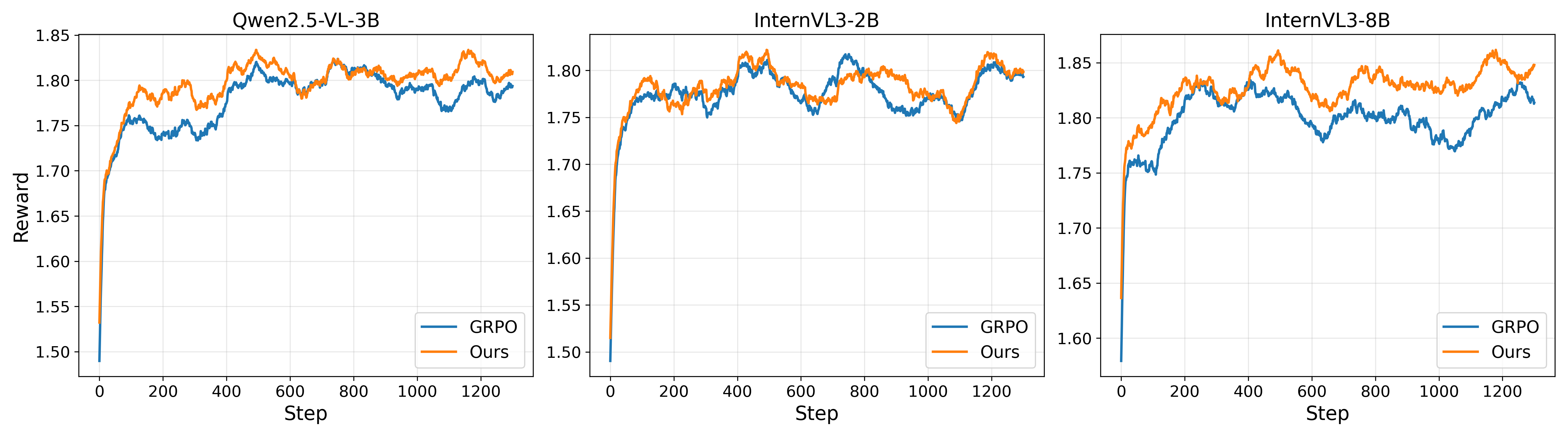}

   \caption{Comparison of reward's trend between our proposed method and the GRPO method during training. The lines in the figure are smoothed.}
   \label{fig:trend_reward}
\end{figure*}
\Cref{fig:trend_reward} compares the reward trends of our proposed EC-GRPO method and the GRPO baseline during training on Qwen2.5-VL-3B, InternVL3-2B, and InternVL3-8B. Across all models, ECVGPO consistently achieves higher average rewards and exhibits smoother convergence compared to GRPO. This indicates that our method enables more effective policy updates and better balances exploration and exploitation during training.
\label{sec:appendix:ablation}
\paragraph{Comparison with different training data}
\begin{table}[!h]
\centering
  \setlength\tabcolsep{5pt}

  \scalebox{0.73}{
\begin{tabular}{lcccc}
    \toprule
    Setting  & RefCOCO$_{val}$      &RefCOCO+$_{val}$ & RefCOCOg$_{val}$ & LISA \\
    \midrule
      Base Model & \textbf{93.00} & \textbf{88.40} & 89.35  & 0.84   \\
    \midrule
     w/ RefGTA   & 27.90   & 26.65 & 30.95  &4.46  \\
     w/ RefCOCO   & 92.15$_{\pm0.15}$   & 88.13$_{\pm0.33}$ & \textbf{89.43}$_{\pm0.41}$   & \textbf{81.14}$_{\pm0.45}$\\

 \hline
\end{tabular}
}
    \caption{Result of different training data. The best performance is \textbf{bolded}. The experiments are conducted on the InternVL-8B model.
}
      \label{tab:data_result}
\end{table}
we experiment with different types of training data and find that RefCOCO, a general visual grounding dataset covering a wide range of scenarios, yields the best performance, as shown in~\cref{tab:data_result}. In contrast, RefGTA, which contains limited and less diverse scenarios, often leads to training instability or even collapse.
\paragraph{Comparison with different temperture}
As shown in~\cref{tab:temperature_result}, we evaluate the impact of different sampling temperatures on model performance. Our method consistently outperforms the GRPO baseline across all temperature settings, demonstrating strong robustness and generalization. Specifically, at the standard temperature of 1.0, our method achieves better results, while maintaining competitive performance when the temperature is varied to 0.8 or 1.2.
\paragraph{Comparison with different rollout}
As shown in Table~\ref{tab:rollout_result}, we investigate the effect of different rollout numbers during training. To prevent model collapse, a smaller intervention (\(r_0\)=75) was used for rollout=16. Our method consistently surpasses the GRPO baseline across all rollout settings, indicating its robustness and stability. Notably, the improvement is most evident when the rollout is set to 8, where our approach achieves the best overall performance on all benchmarks.
\begin{table*}[!ht]
\centering
  \setlength\tabcolsep{5pt}

  \scalebox{1.0}{
\begin{tabular}{lccccc}
    \toprule
    Setting  & RefCOCO$_{val}$      &RefCOCO+$_{val}$ & RefCOCOg$_{val}$ & LISA & RefGTA\\
        \midrule
        \rowcolor[gray]{0.9}\multicolumn{6}{c}{\textbf{temperature = 1.0}} \\

      GRPO & 91.65$_{\pm0.65}$ & 87.33$_{\pm0.52}$ & 88.97$_{\pm0.46}$  & 78.78$_{\pm1.45}$  &48.80$_{\pm6.30}$\\
     w/ Ours   & \textbf{92.15}$_{\pm0.15}$   & \textbf{88.13}$_{\pm0.33}$ & \textbf{89.43}$_{\pm0.41}$ & \textbf{81.14}$_{\pm0.45}$ &\textbf{51.29}$_{\pm6.24}$\\
     
        \rowcolor[gray]{0.9}\multicolumn{6}{c}{\textbf{temperature = 0.8}} \\

      GRPO & 91.10$_{\pm0.43}$ & 87.32$_{\pm0.26}$ & 89.05$_{\pm0.25}$  & 79.02$_{\pm0.77}$   &\textbf{47.67}$_{\pm6.21}$\\
     w/ Ours   & \textbf{92.28}$_{\pm0.29}$   & \textbf{87.86}$_{\pm0.30}$ & \textbf{89.18}$_{\pm0.32}$ &\textbf{80.02}$_{\pm0.42}$ &45.28$_{\pm3.59}$\\
        \rowcolor[gray]{0.9}\multicolumn{6}{c}{\textbf{temperature = 1.2}} \\

      GRPO & 92.13$_{\pm0.23}$ & 87.36$_{\pm0.45}$ & 88.95$_{\pm0.51}$  & 77.85$_{\pm1.20}$   &29.79$_{\pm2.37}$\\
     w/ Ours   & \textbf{92.13}$_{\pm0.27}$    & \textbf{87.53}$_{\pm0.25}$ & \textbf{89.22}$_{\pm0.42}$& \textbf{78.37} $_{\pm1.93}$  &\textbf{35.68}$_{\pm4.06}$\\

 \hline
\end{tabular}
}
    \caption{Result of different temperature of our method. The best performance is \textbf{bolded}. The experiments are conducted on the InternVL-8B model. We report the mean and standard deviation over four runs.
}
      \label{tab:temperature_result}
\end{table*}
\begin{table*}[!ht]
\centering
  \setlength\tabcolsep{5pt}

  \scalebox{1.0}{
\begin{tabular}{lccccc}
    \toprule
    Setting  & RefCOCO$_{val}$      &RefCOCO+$_{val}$ & RefCOCOg$_{val}$ & LISA & RefGTA\\
        \midrule
        \rowcolor[gray]{0.9}\multicolumn{6}{c}{\textbf{rollout = 8}} \\

      GRPO & 91.65$_{\pm0.65}$ & 87.33$_{\pm0.52}$ & 88.97$_{\pm0.46}$  & 78.78$_{\pm1.45}$  &48.80$_{\pm6.30}$\\
     w/ Ours   & \textbf{92.15}$_{\pm0.15}$   & \textbf{88.13}$_{\pm0.33}$ & \textbf{89.43}$_{\pm0.41}$ & \textbf{81.14}$_{\pm0.45}$ &\textbf{51.29}$_{\pm6.24}$\\
     
        \rowcolor[gray]{0.9}\multicolumn{6}{c}{\textbf{rollout = 4}} \\
        GRPO & 91.68$_{\pm0.14}$ & \textbf{87.83}$_{\pm0.16}$ & 89.06$_{\pm0.27}$  & 78.35$_{\pm1.27}$   &37.87$_{\pm7.09}$\\
     w/ Ours   & \textbf{91.74}$_{\pm0.13}$    & 87.61$_{\pm0.33}$ & \textbf{89.16}$_{\pm0.27}$& \textbf{79.37} $_{\pm2.34}$  &\textbf{47.29$_{\pm13.69}$}\\

        \rowcolor[gray]{0.9}\multicolumn{6}{c}{\textbf{rollout = 16}} \\
         GRPO & 92.35$_{\pm0.39}$ & \textbf{88.38}$_{\pm0.36}$ & \textbf{89.56}$_{\pm0.31}$  & 78.64$_{\pm0.43}$   & 41.67$_{\pm5.91}$\\
     w/ Ours   & \textbf{92.54}$_{\pm0.35}$   & 88.21$_{\pm0.25}$ & 89.48$_{\pm0.36}$ &\textbf{79.71}$_{\pm0.65}$ &\textbf{52.58}$_{\pm4.71}$\\

 \hline
\end{tabular}
}
    \caption{Result of different rollouts of our method. The best performance is \textbf{bolded}. The experiments are conducted on the InternVL-8B model. We report the mean and standard deviation over four runs.
}
      \label{tab:rollout_result}
\end{table*}
\section{Limitation}
While ECVGPO demonstrates strong performance and stability improvements, several limitations remain. First, due to the intrinsic uncertainty caused by high entropy in visual grounding, the evaluation results
 can still exhibit noticeable fluctuations, which may pose challenges for reproducibility. Second, the method involves several hyperparameters that require careful tuning, making the optimization process time-consuming and somewhat inconvenient in practice. Finally, although our experiments cover multiple benchmarks, further validation on broader scenarios and diverse multimodal tasks is needed to fully assess the generality and robustness of ECVGPO.
\section{Case Study}
\begin{figure*}[ht!]
  \centering
   \includegraphics[width=0.72\linewidth]{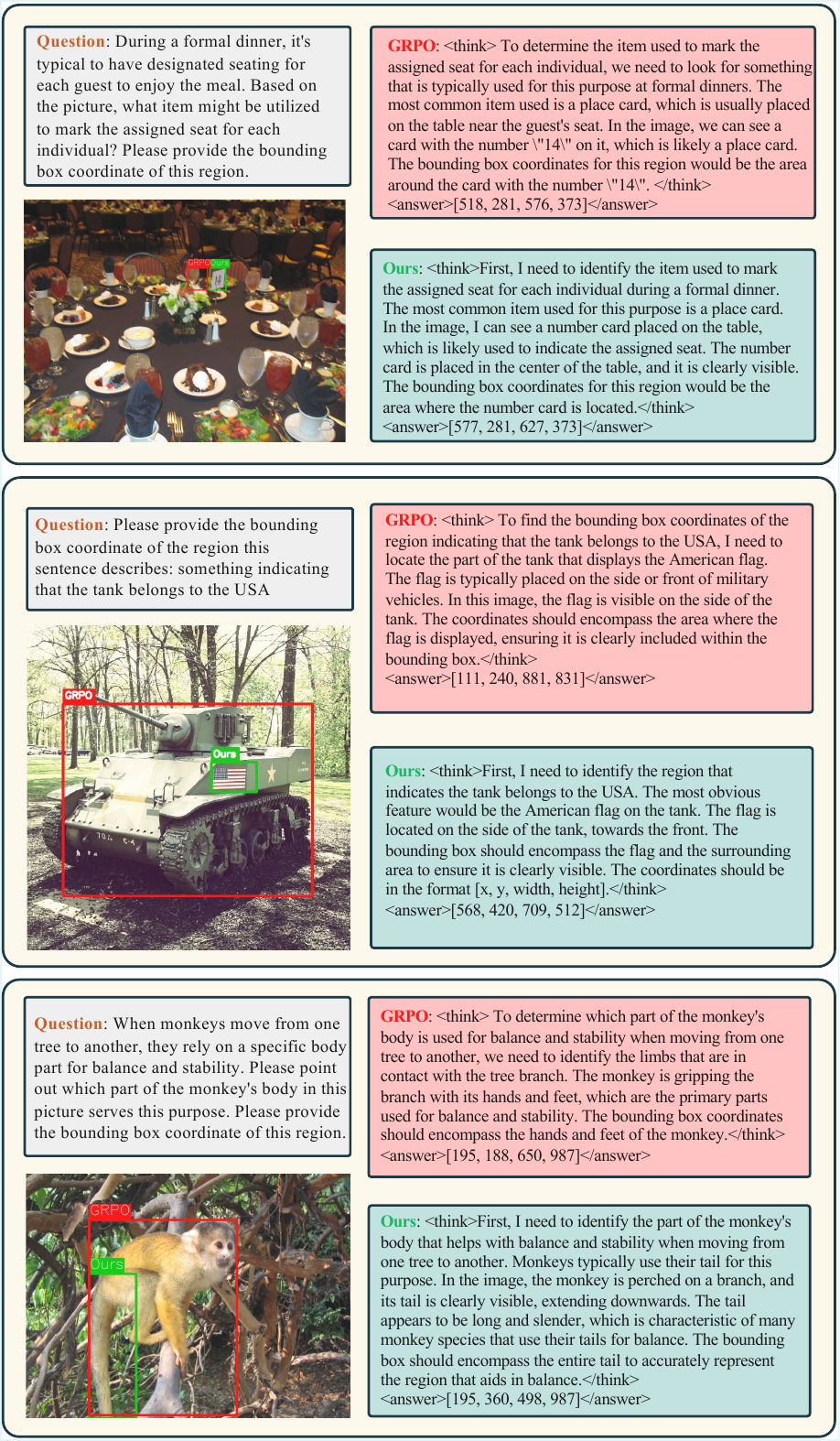}

   \caption{Comparison between our proposed method and the GRPO method in some cases.}
   \label{fig:case_study}
\end{figure*}

\Cref{fig:case_study} presents a case-by-case comparison between our proposed method and the GRPO method. The results demonstrate that our method significantly improves the model's visual grounding 
capabilities. 



\end{document}